# Pre-processing in AI based prediction of QSARs

Om Prasad Patri, Amit Kumar Mishra
Department of Computer Science & Engineering, and
Department of Electronics and Communication Engineering,
Indian Institute of Technology Guwahati, India
E-mail: {o.patri, amishra}@iitg.ernet.in

**ABSTRACT**

Machine learning, data mining and artificial intelligence (AI) based methods have been used to determine the relations between chemical structure and biological activity, called quantitative structure activity relationships (QSARs) for the compounds. Pre-processing of the dataset, which includes the mapping from a large number of molecular descriptors in the original high dimensional space to a small number of components in the lower dimensional space while retaining the features of the original data, is the first step in this process. A common practice is to use a mapping method for a dataset without prior analysis. This pre-analysis has been stressed in our work by applying it to two important classes of QSAR prediction problems: drug design (predicting anti-HIV-1 activity) and predictive toxicology (estimating hepatocarcinogenicity of chemicals). We apply one linear and two nonlinear mapping methods on each of the datasets. Based on this analysis, we conclude the nature of the inherent relationships between the elements of each dataset, and hence, the mapping method best suited for it. We also show that proper preprocessing can help us in choosing the right feature extraction tool as well as give an insight about the type of classifier pertinent for the given problem.

**Keywords**

QSAR, machine learning, predictive toxicology, clustering, drug design, chemoinformatics

## 1. INTRODUCTION

Artificial intelligence (AI) techniques have been increasingly used as helpful tools in drug design [8] and predictive toxicology [11,13] for determining relations between structural features of a molecule and its biological activity, termed as quantitative structure-activity relationships (QSAR). These techniques include approaches based on statistical and machine learning, pattern recognition, clustering, similarity-based methods, as well as biologically motivated approaches, such as neural networks [5,7,10,15-16], evolutionary computing or fuzzy modeling [9], collectively described as "computational intelligence" [23,24]. Applications of AI methods involve selection of relevant information, data visualization, classification and regression, optimization and prediction.

A large number of potential drug candidates fail due to poor absorption, distribution, metabolism, elimination or toxicity (ADMET) properties, thus giving rise to one of the most costly problems in drug design. Recent studies have shown ADMET problems to be the reason for the failure of 60% of drug candidates [4]. Suitable computational methods can greatly help in the drug design process by predicting activity of compounds before they are actually manufactured.

It is often easier, and more feasible, to visualize one-dimensional or two-dimensional data as compared to data in higher dimensions. Mapping or 'feature extraction' tools are used to extract appropriate features of the higher dimensional data so that it can be represented in lower dimensions while retaining the inherent features of the original high dimension data. In this work, we have considered one linear mapping method – principal component analysis (PCA), [17,18] and two nonlinear mapping methods – nonlinear PCA [25] and sammon's nonlinear mapping (NLM) [22] for pre-analysis of the compounds in the two datasets. Determining the best mapping tool for a dataset is a very important step which is usually ignored and we show that the best mapping method for a certain dataset depends on the structural relationships between the elements of the dataset.

Pre-processing, or suitably extracting features from the data to make it ready for the AI system, is a major step in any predictive QSAR problem. It includes removal of duplicate compounds, standardization of the data and most importantly, dimensionality reduction. In this paper we concentrate on two major classes of AI applications: prediction and classification. We apply and show the importance of choosing proper pre-processing techniques on two different sets of compounds which have significant implications in medical science: (1) human immunodeficiency virus-1 (HIV-1), with anti-HIV-1 activity of the compounds as the endpoint and (2) carcinogenicity, specifically hepatocarcinogenic toxicity (carcinogenic potential for liver cancer) with carcinogenic potency of the compounds as the endpoint.

The first dataset used in our work is a group of 80 compounds first reported in [2] and used by [1,5-7]. This series of compounds, contained in the group of 1-[2-hydroxyethoxy-methyl]-6-(phenylthio)thymine], or HEPT derivatives, have been shown to be potent inhibitors of the reverse transcriptase (RT) enzyme of HIV-1. RT is a prime target for antiviral therapy against acquired immune deficiency syndrome (AIDS) [1]. The endpoint used is $pIC_{50}$ values[1]. Previous work on this dataset of compounds includes linear methods [1,6,10] as well as nonlinear methods like artificial neural networks [5,7]. Pre-processing of the dataset is an important step in all of the above mentioned works.

---

[1]$pIC_{50}$ is the logarithm of reciprocal molar concentration required to achieve 50% protection of MT-4 cells against the cytopathic effect of HIV-1, from [2]. Higher the $pIC_{50}$ value, greater the anti-HIV-1 activity of the compound.

The second dataset we use is obtained from the carcinogenic potency database (CPDB) [3]. Predicting carcinogenicity and mutagenicity and determining QSARs for them has been researched actively [4,12-16]. Effective prediction methods help identify potential carcinogens and thus reduce attrition rates due to ADMET failures as well as rampant animal testing. We use a dataset of 55 potential hepatocarcinogenic compounds from the CPDB database which are also reported by [4]. The endpoint used is the activity score from the CPDB database which is directly proportional to carcinogenic potency.

Almost all methods used for toxicity prediction use some method for pre-processing of the initial dataset. A very common practice is to use a supposedly 'better' mapping method in the pre-processing step without proper analysis. For instance, a nonlinear feature extraction method is often considered to be 'superior' to a linear mapping method. However, this need not be the case always. The type of mapping best suited for the dataset depends on the nature of the dataset and the relationships between the elements of the dataset.

Also, the type of classifier, i.e. linear or quadratic etc, most apt for division of the data into clusters (carcinogenic and non-carcinogenic or toxic and non-toxic), can be selected by inspection of the plots between the components of the lower dimensional space. We show this by applying both linear and nonlinear mapping methods to the datasets mentioned above. Yet another advantage of this analysis is that if we know the mapping method and the classifier, we can decide the most suitable descriptors for the dataset from these plots. Through this work, we aim to show the importance of proper analysis before choosing a certain pre-processing method for similar problems.

The organization of the paper is as follows. Section 2 reviews the theory about the three mapping methods - PCA, nonlinear PCA and Sammon's NLM which have been used in this work. Section 3 specifies the detailed procedure followed by us in this work for the pre-processing analysis. Section 4 states the observations and results and depicts the plots obtained for the datasets while Section 5 sums up the paper.

## 2. THEORY

Choosing proper components in the lower dimensional space (to represent higher dimensional data) is of prime importance in the pre-processing step. Mapping tools used to solve this problem may be linear like principal component analysis (PCA) and linear discriminant analysis (LDA) or non-linear like nonlinear principal component analysis (NLPCA), kernel PCA and sammon's nonlinear mapping (NLM) algorithm.

Principal component analysis (PCA), is a linear mapping method for dimensionality reduction which has a wide range of applications. [17-18] The dimensionality reduction is done by considering eigen vectors of the original data from the covariance matrix, retaining the most significant eigen vectors and using them to construct linear principal components (PCs) for the lower dimensional space.

Nonlinear principal component analysis (NLPCA) is a nonlinear generalization of PCA [25]. It generalizes the principal components from straight lines to curves (nonlinear). Nonlinear PCA can be achieved by using a neural network with an autoassociative architecture also known as autoencoder, bottleneck or sandglass type network.

Sammon's nonlinear mapping algorithm (NLM) [22] tries to preserve the inherent structure of the vectors of the original higher dimension vectors in the output lower dimension vectors. This is done by fitting the points in the lower dimension space such that their interpoint distances approximate the corresponding interpoint distances in the higher dimension space.

## 3. PROCEDURE

The first dataset of 80 compounds, as mentioned above, was used from [1]. The other carcinogenicity dataset was obtained from the CPDB database [3]. The Carcinogenic Potency Database (CPDB) is a widely used international resource of the results of 6540 long-term animal cancer tests on 1547 chemicals. For best results and maximizing homogeneity of the dataset, only compounds with hepatocarcinogenic potential and only data corresponding to a single species 'mouse' and a single sex 'female' were considered. The two datasets are completely different from each other and, as we show, of different structures. This shows that the methods used for them can be easily adapted for a completely new dataset with different relationships between its elements.

For the HEPT derivatives dataset, the 10 descriptors used for each chemical were used as it is from [1]. This set of descriptors has been shown to encapsulate well most of the chemical properties of the dataset.

For the carcinogenicity dataset, we first calculated a large number of descriptors, 23 for each compound in the dataset. These descriptors included essential global molecular features like molecular weight, number of atoms, octanol-water coefficient, dipole moment and polarizability and shape descriptors like eccentricity, asphericity, moment of inertia and radius of gyration of the molecules. A complete list of the 23 descriptors is given in Table 1. The descriptors were normalized such that the whole dataset had a mean of zero and standard deviation of one. This helped to reduce large differences in the absolute values of data while keeping the relative differences intact, thus helping in better visualization and plotting of the data. The software Adriana.Code [20] was used to calculate the descriptors. The code for sammon's mapping algorithm was obtained from [21] and the nonlinear PCA toolbox from [19].

**Table 1:** The 23 molecular descriptors used in pre-processing of the carcinogenicity dataset

| Notation | Explanation |
|---|---|
| Weight | Molecular weight |
| HDon | No. of H-bond donors |
| HAcc | No. of H-bond acceptors |
| XlogP | Octanol-water distribution coefficient |
| TPSA | Topological polar surface area |
| Polariz | Mean molecular polarizability |
| Dipole | Molecular dipole moment |
| LogS | Aqueous solubility |
| NRotBond | No. of rotatable bonds |
| NVRo5 | No. of Lipinski's rule of 5 violations |
| NVERo5 | Extended Lipinski's rule of 5 violations |
| NAtoms | Number of atoms |
| NStereo | No. of tetrahedral stereocenters |
| Complexity | Molecular complexity |
| RComplexity | Ring complexity |
| Diameter | Molecular diameter |
| InertiaX, InertiaY, InertiaZ | $1^{st}$, $2^{nd}$ and $3^{rd}$ PC of moment of inertia |
| Span | Molecular span |
| RGyr | Molecular radius of gyration |
| Eccentric | Molecular eccentricity |
| Aspheric | Molecular asphericity |

The initial 10 or 23-dimensional data was reduced to one or two dimensions for better visualization. In two-dimensional space, the two principal components were plotted against each other while in one-dimensional space, the principal component was plotted against the index number of the compound. The datasets were divided into two sets each (depending upon their biological activity) for visualization of clusters. For the anti-HIV dataset, the compounds with $pIC_{50}$ greater than 6 were considered to highly inhibit the HIV-1 RT, and thus show better anti-HIV activity (*active*) while the others were labeled *inactive*. For the carcinogenicity dataset, compounds with an activity score greater than 29 (which was the mean of the dataset) were considered to be *toxic* (potential hepatocarcinogens) and the others as *non-toxic* (safe compounds). This caused even distribution of the data into *toxic* and *non-toxic* divisions.

## 4. OBSERVATIONS AND RESULTS

The 1-dimensional and 2-dimensional plots were inspected for clusters of *toxic/non-toxic* and *active/inactive* compounds based on the above criteria.

A mapping method, which shows better division or clustering of the compounds into *toxic/non-toxic* or *active/inactive*, is a better mapping method for the particular dataset. Also, these plots can be used to determine the right type of classifier i.e. linear, quadratic etc best suited for the method.

The results of applying the mapping methods and the clustering analysis to the anti-HIV (HEPT compounds) dataset are shown in Figures 1-6. The red crosses indicate *active* compounds with potential anti-HIV activity and the blue asterisks point to *inactive* compounds with lesser or no anti-HIV activity.

For the HEPT compounds in the anti-HIV activity dataset, we find that PCA gives the best clustering in both one-dimensional and two-dimensional output space as compared to the two nonlinear methods. This shows the inherent linear nature of the dataset. Sammon's NLM also gives pretty good results in both one and two-dimensional output spaces. However, NLPCA does not give a good clustering in either. Also, we can observe that for PCA, a simple linear classifier will be adequate for division of the dataset into the two clusters: *active* and *inactive* with regards to their anti-HIV-1 activity.

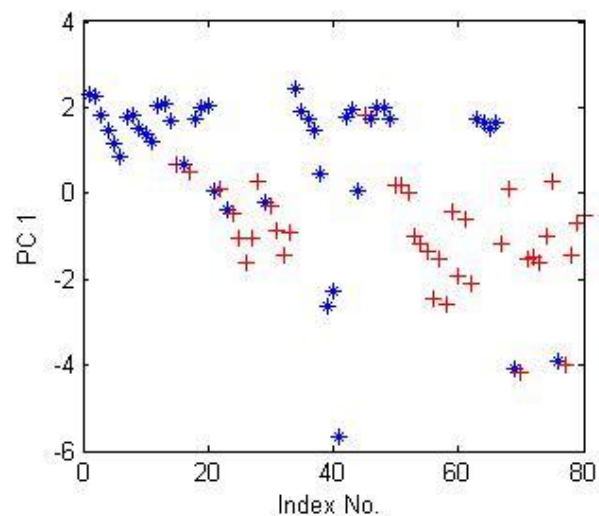

**Figure 1**: PCA (1-dimension) on HIV dataset

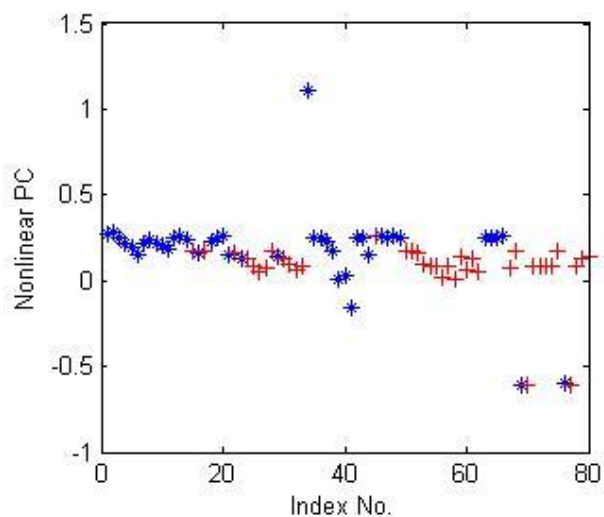

**Figure 2**: Nonlinear PCA (1-dimension) on HIV dataset

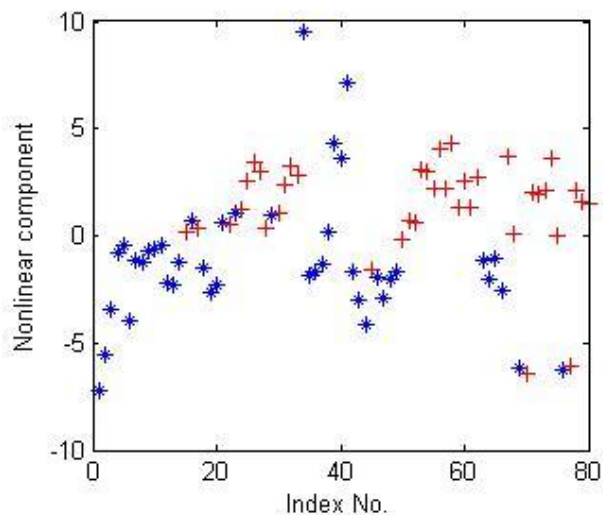

**Figure 3**: Sammon's nonlinear mapping (1-dimensional output space) on HIV dataset

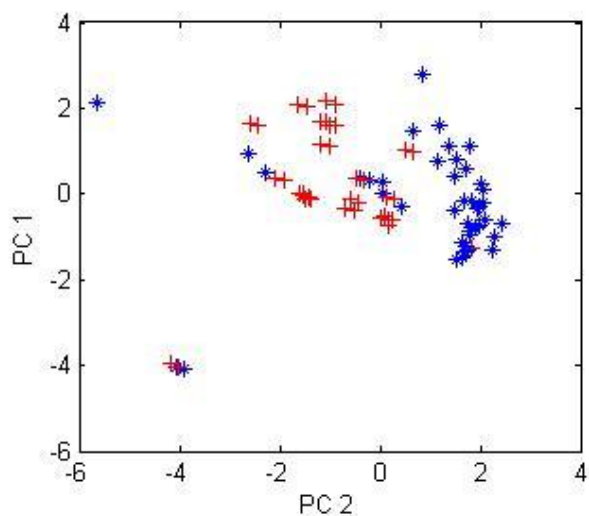

**Figure 4**: PCA (2-dimensions) on HIV dataset

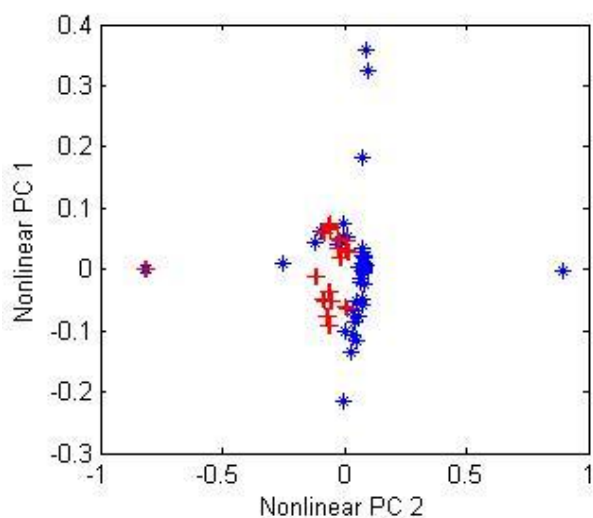

**Figure 5**: Nonlinear PCA (2-dimensions) on HIV dataset

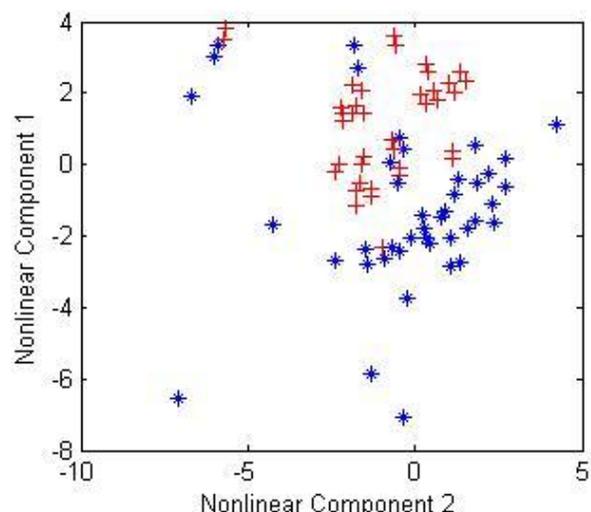

**Figure 6**: Sammon's nonlinear mapping (2-dimensional output space) on HIV dataset

The observations on pre-processing and clustering of the carcinogenicity dataset are shown in Figures 7-12. The red crosses indicate *toxic* compounds which are probable hepatocarcinogens and the blue asterisks indicate *inactive* compounds which do not show hepatocarcinogenic behavior.

From the hepatocarcinogenicity plots, it can be noted that all the three methods give a good clustering in one-dimensional output space. Also, a linear classifier will be sufficient. In case of the two dimensional output space, none of the three mapping methods give a clear distribution into two clusters. However, sammon's NLM gives a clear separation if we consider more than one cluster of red crosses and the blue asterisks i.e. *toxic* and *non-toxic* compounds. Further, it can be observed that to separate the two clusters, a simple linear classifier will not suffice in this case.

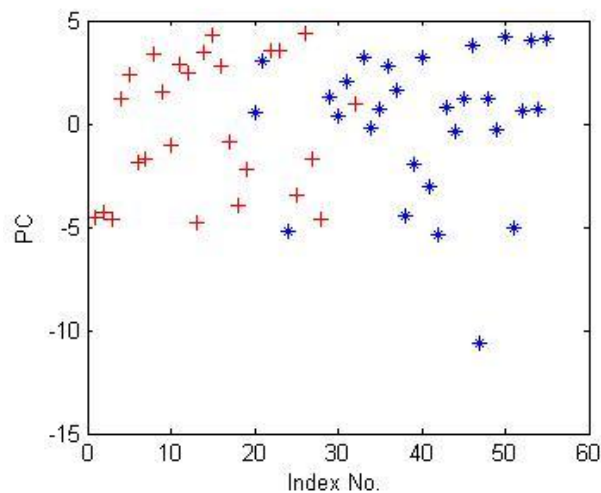

**Figure 7**: PCA (1-dimension) applied to hepatocarcinogenicity dataset

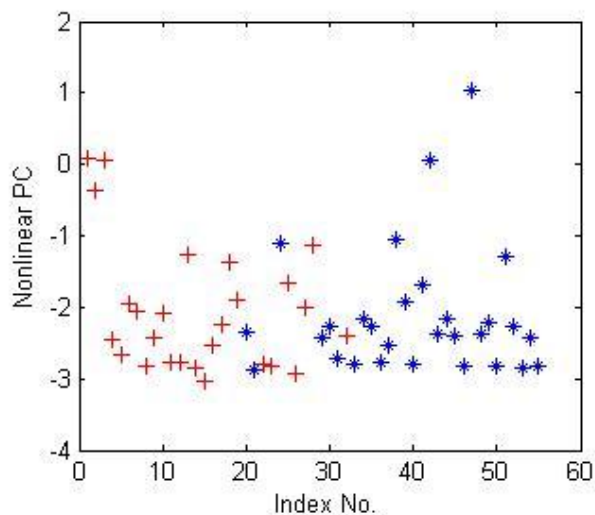

**Figure 8**: Nonlinear PCA (1-dimension) applied to hepatocarcinogenicity dataset

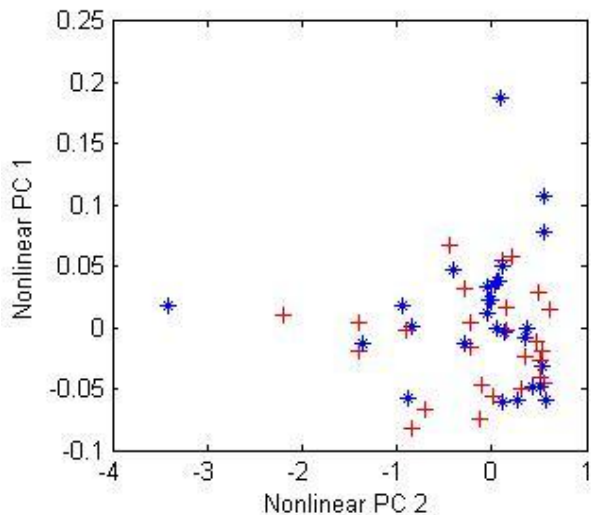

**Figure 11**: Nonlinear PCA (2-dimensions) applied to hepatocarcinogenicity dataset

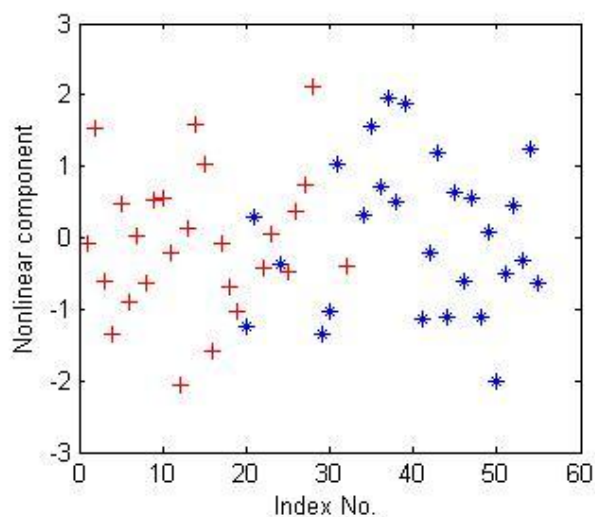

**Figure 9**: Sammon's nonlinear mapping (1-dimensional output space) applied on liver carcinogenicity dataset

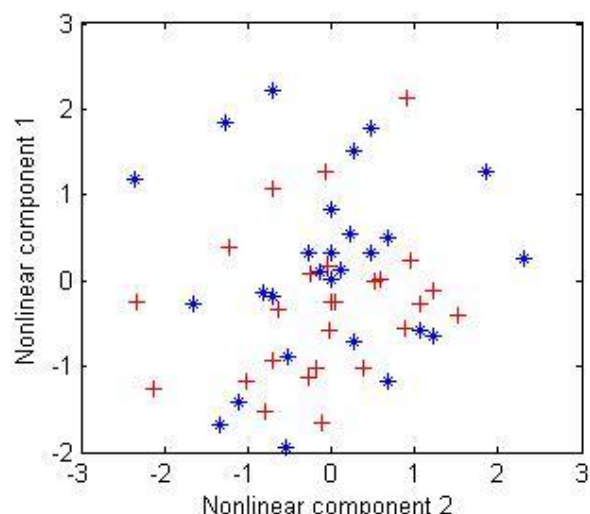

**Figure 12**: Sammon's nonlinear mapping (2-dimensional output space) on hepatocarcinogenicity dataset

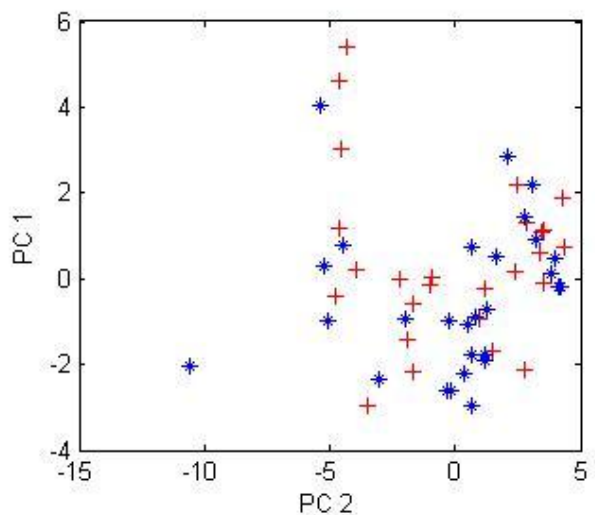

**Figure 10**: PCA (2-dimensions) applied to hepatocarcinogenicity dataset

## 5. CONCLUSION

All artificial intelligence methods for QSAR determination use some form of pre-processing. Through this work, our objective has been to depict the importance of the first and important step of pre-processing of datasets for estimating biological activity, as in drug design as well as predictive toxicology. A nonlinear mapping method need not be better than a linear one as exhibited by the first dataset in our work. In fact, the suitability depends upon the intrinsic relationships between the elements of the dataset. Proper pre-processing of the dataset is a must for every such problem and pre-analysis before choosing the type of mapping method can lead to better efficiency in the prediction of toxicity as well as other similar problems. Also, insights about the appropriate type of classifier can be obtained from the plots between the principal components.